\documentclass[conference]{IEEEtran}
\IEEEoverridecommandlockouts

\usepackage{cite}
\usepackage{amsmath,amssymb,amsfonts}
\usepackage{algorithmic}
\usepackage{graphicx}
\usepackage{textcomp}
\usepackage{xcolor}
\usepackage[colorinlistoftodos]{todonotes}

\usepackage{layouts}

\usepackage{tikz}
\usetikzlibrary{decorations.pathreplacing, positioning, shapes, arrows.meta}
\usepackage[caption=false,font=footnotesize]{subfig}
\usepackage{algorithm}
\usepackage{algorithmic}
\usepackage{multirow}
\RequirePackage{rsfso}
\usepackage{booktabs}
\usepackage{balance}

\definecolor{linkblue}{RGB}{50,100,170} 
\usepackage{hyperref}
\hypersetup{
    pdftitle = {Tile},
    pdfauthor = {Author},
    pdfborder={0 0 0},
    colorlinks=true,
    urlcolor=linkblue,
    linkcolor=linkblue,
    citecolor=linkblue,
    bookmarks=true,
    bookmarksnumbered=true
}

\def\BibTeX{{\rm B\kern-.05em{\sc i\kern-.025em b}\kern-.08em
        T\kern-.1667em\lower.7ex\hbox{E}\kern-.125emX}}

\newif\ifanonymous
\anonymousfalse

\makeatletter
\newcommand\thefontsize[1]{{#1: \f@size pt\par}}
\makeatother

\begin{document}
    
    \title{Fruit Picker Activity Recognition with Wearable Sensors and Machine Learning
        \ifanonymous
        \thanks{This work is a product of a collaboration with \textit{Anonymous}, who provided the dataset, photographs, and the context to the problem.}
        \else
        \thanks{This work is a product of a collaboration with Grow Logic, who provided the dataset, photographs, and the context to the problem. The work was supported by the Systems program in Agriculture \& Food, CSIRO.}
        \fi
    }
    
    \ifanonymous
    \author{\IEEEauthorblockN{Anonymous Authors}}
    \else
    \author{
        \IEEEauthorblockN{Joel Janek Dabrowski$^*$}
        \IEEEauthorblockA{
            \textit{Data61, CSIRO} \\
            Brisbane, Australia \\
            joel.dabrowski@data61.csiro.au}
        \and
        \IEEEauthorblockN{Ashfaqur Rahman}
        \IEEEauthorblockA{
            \textit{Data61, CSIRO} \\
            Hobart, Australia \\
            ashfaqur.rahman@data61.csiro.au}
    }
    \fi
    
    \maketitle
    
    \begin{abstract}
        In this paper we present a novel application of detecting fruit picker activities based on time series data generated from wearable sensors. During harvesting, fruit pickers pick fruit into wearable bags and empty these bags into harvesting bins located in the orchard. Once full, these bins are quickly transported to a cooled pack house to improve the shelf life of picked fruits. For farmers and managers, the knowledge of when a picker bag is emptied is important for managing harvesting bins more effectively to minimise the time the picked fruit is left out in the heat (resulting in reduced shelf life). We propose a means to detect these bag-emptying events using human activity recognition with wearable sensors and machine learning methods. We develop a semi-supervised approach to labelling the data. A feature-based machine learning ensemble model and a deep recurrent convolutional neural network are developed and tested on a real-world dataset. When compared, the neural network achieves 86\% detection accuracy.
    \end{abstract}
    

%


\section{Introduction}

It is estimated that up to 25\% of all fruit and vegetable produce go to waste before ever leaving the farm, costing Australian farmers $\$2.84$ billion annually \cite{Larissa2020Food}. Waste may be caused by pests, disease, weather events, or damage inflicted during harvesting. In harvesting, fruit are picked and placed in harvesting bins located throughout the paddocks or orchards, and transported to a packing house for processing prior to market distribution. The picked fruit are spoiled when they are left in the harvesting bins exposed to heat for extended periods of time. Farmers and managers must thus ensure that the pickers fill the bins quickly and that the bins are transported to the cooled packing house as soon as they are full.

In general, the machine learning paradigm provides promising approaches to address challenges in harvesting, however examples of such applications are scarce in the literature \cite{altalak2022smart}. This study contributes to this gap in the literature and provides a means to monitor fruit pickers and the filling of the harvesting bins to allow for more effective management of the harvesting process.

\begin{figure}[!t]
    \centering
    \input{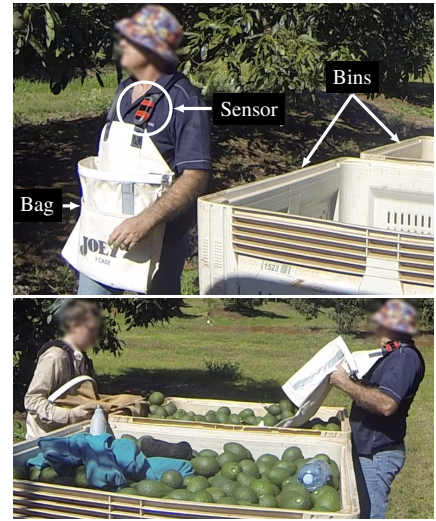}
    \caption{\textit{Top:} Photograph of a fruit picker, a picker bag, sensor, and harvesting bin. The bin is mounted on a trailer, which is typically pulled by a tractor. There are typically many bins scattered across the orchard. \textit{Bottom:} Photograph of bag-emptying events where two fruit pickers empty the bags into a bin.}
    \label{fig:fruit_picker}
\end{figure}

As illustrated in \figurename{~\ref{fig:fruit_picker}}, fruit pickers are equipped with a fruit picker bag. Fruit are picked and placed into these bags, and once the bag is full, the fruit is emptied into a harvesting bin. This is referred to as a \textit{bag-emptying event}. Detecting bag-emptying events provides the means to simultaneously monitor the fruit picker productivity and bin levels. The bag-emptying event rate is a direct measurement of the fruit picker productivity. Furthermore, given the known capacity of a picker bag and the knowledge of which bin the fruit is emptied into, bag-emptying events provide a measure of the filling rate of the bins. With this knowledge, farmers are able to more effectively manage pickers and bins.

Given recent success in the application of deep and machine learning approaches to human activity recognition \cite{zhang2022deep}, we propose a novel application of these approaches to detect bag-emptying events using wearable sensors. A wearable sensor is placed on a picker's bag strap as illustrated in \figurename{~\ref{fig:fruit_picker}} to measure the picker's movements and their proximity to the harvest bins. Given the sensor measurements, bag-emptying events are detected using machine learning models. We consider two models: a feature-based machine learning ensemble classifier and a deep Recurrent Convolutional Neural Network (RCNN). The models are compared on a dataset collected during avocado harvesting.

In addition to bag-emptying event detection, we address a data labelling problem where the duration of bag-emptying events are unknown. For this, the K-means clustering algorithm is used to perform semi-supervised labelling of the data. Approximate times of when the events occurred and several features of the data are used to learn the duration of bag-emptying events.

To our knowledge, fruit picker activity recognition based on time series data from wearable sensors is a novel application. The main contributions of this study include: (1) we use a wearable accelerometer sensor for measuring fruit picker activity, (2) we used machine learning to understand fruit picker productivity,  and (3) we develop a new methodology that uses a combination of semi-supervised labelling and supervised learning to detect bag-emptying events. The key advantages of our approach is it does not interfere with the harvesting process and it is autonomous.

This article begins with a discussion on related work in Section \ref{sec:har}. The dataset and data labelling approaches are discussed in Sections \ref{sec:dataset} and \ref{sec:bagDropLabelling} respectively. The models are described in Section \ref{sec:models} and our methodology is provided in Section \ref{sec:methodology}. Results are presented in Section \ref{sec:results} and the article is concluded in Section \ref{sec:conclusion}.


\section{Related Work}
\label{sec:har}

\subsection{Machine Learning in Agriculture}

Machine learning has been applied to various agricultural problems and several literature surveys have been produced \cite{altalak2022smart,benos2021machine,sharma2020systematic}. Most applications relate to crop management and include yield prediction, disease detection, weed detection, crop quality prediction, and water management. Literature on the application of machine learning approaches specifically to harvesting and the harvesting process are scarce.

\subsection{Fruit Picker Productivity}

In relation to fruit picker activity recognition, several studies have considered tracking pickers for yield mapping, e.g., see \cite{Khosro2018Development, whitney2001dgps, thomas1999development, Pelletier1999Development, miller1999evaluation}. Tracking pickers can be challenging due to Global Positioning System (GPS) signal losses through foliage. Various alternatives to GPS have been thus been proposed \cite{Ampatzidis2011wearable,Arikapudi2016Estimation}. However, even accurate picker tracking does not necessarily provide any direct information on the picker's activity. In our work, we are directly measuring the fruit picker's activity, which to our knowledge has not been considered in the literature before.

A more direct approach to measuring picker productivity is to have a fixed platform located at the bin containing some form of digital scale and a device to identify the picker (such as RFID or a bar code scanner) \cite{Ampatzidis2016Cloud, Ampatzidis2013Portable, Ampatzidis2012Development, Ampatzidis2009Field}. The picker identifies them self, weighs the picked fruit, and releases the weighed fruit into the bin. Drawbacks of this approach include the additional time that is required for weighing bags and the additional supervisors required to ensure that the weighing processes is being conducted correctly. Our approach does not interfere with the harvesting process and does not require additional supervisors. Furthermore, our approach considers time-series data rather than data at a single point in time.

\subsection{Human Activity Recognition and Machine Learning}

Detecting fruit picker bag-emptying events can be considered as a Human Activity Recognition (HAR) problem. Generally, HAR involves using some form of classifier to predict an activity given data from a sensor that directly monitors human movement, such as wearable sensors. Surveys on HAR using wearable sensors \cite{Lara2013Survey} deep learning approaches to HAR \cite{zhang2022deep} have been conducted. Applications with wearable sensors in agriculture include human–robot interaction \cite{anagnostis2021human} and the assessment of vibration risk with agricultural machinery \cite{aiello2022worker}. To our knowledge HAR has not been applied to monitor fruit pickers.

A wide range of wearable sensors exist \cite{Lara2013Survey} including: accelerometers, global positioning systems (GPS), radio frequency identification (RFID), environmental sensors (such as temperature), and physiological sensors (such as heart rate monitors). A survey has been conducted on sensor positioning on the body \cite{Atallah2011Sensor}. Sensor positions may include the waist, arms, wrist, ankle, and the torso. The chest is suggested to be the preferred location for medium-level activities such as walking and house-work. The actions in such activities are similar to that of fruit picking. The chest was thus chosen as the location for this study.

Many HAR models begin by processing the accelerometer data using a sliding window \cite{Lara2013Survey}. Various features are extracted from the data in the window as it is slid across the dataset. Features may include mean, standard deviation, minimum, maximum, energy, main frequency component, root mean square of the derivative, and correlation between axes \cite{Atallah2011Sensor, Ravi2005Activity}. The features are fed into a classifier, which classifies the activity type. Various classifiers such as decision trees, neural networks, Bayesian models, Markov models, and classifier ensembles have may be considered \cite{Lara2013Survey}. In this study, such a feature-based ensemble classifier is compared with a deep neural network model.

Feature selection and design can be a tedious task that typically requires domain knowledge. Deep learning algorithms are often designed to be end-to-end methods that take the raw input data and output a prediction. Features are learned within the multiple layers of network. A survey of various deep learning architectures that have been applied to sensor-based activity detection problems has been conducted \cite{Wang2018Deep}. These architectures include the convolutional neural network (CNN) and the recurrent neural network (RNN).

The CNN is a neural network that applies a convolution operation on the data presented to its inputs \cite{goodfellow2016deep}. CNNs exploit local interactions in the data and provide scale invariant features \cite{Zeng2014Convolutional}. In wearable sensor HAR problems, several CNN based models have been proposed \cite{Lee2017Human,yang2015deep,Zeng2014Convolutional}.

The RNN is a neural network that is designed for sequential applications \cite{goodfellow2016deep, Graves2012Supervised}. It contains a neural network that is replicated over time where the replications are sequentially connected. Like the CNN, the RNN has been also applied to several HAR problems \cite{Murad2017Deep,Inoue2018Deep,Guan2017Ensembles}.

RNN and CNN models have been compared on HAR tasks \cite{Hammerla2016Deep}. It is found that RNNs perform well for activities where long term dependency is required, such as opening a door. CNNs perform well when long-term dependencies are not required such as gait analysis. Given that RNNs and CNNs each have their own advantages, combining the two architectures may provide a more widely applicable and more powerful model. The combination of the architectures forms a recurrent convolutional neural network (RCNN) and provide a promising framework for HAR \cite{Ordonez2016Deep,Yao2017DeepSense}, and is an approach we consider in this study.

\section{Dataset}
\label{sec:dataset}

\subsection{Sensors and Data Collection}

The Haltian Nexus Prototype sensors were used in the wearable sensor. These sensors comprise a ST LIS2DH accelerometer, a Nordic nRF52832 Wirepas radio gateway module, and a data logger. Accelerometer readings for 3 axes were logged at a frequency of 50Hz with a measurement range of $\pm$4 G and sensitivity of 8 mG. The radio logged its Received Signal Strength Indicator (RSSI) value with respect to a Haltian Thingsee POD2 Prototype node every second. The POD2 node was located at the picker's bin.

\subsection{Avocado Farm Trial Dataset}

A trial was conducted on an avocado farm. The data was acquired for two different pickers over several hours. A sensor was attached to each pickers bag strap and positioned on the picker's chest as illustrated in \figurename{~\ref{fig:fruit_picker}}(a). Bag-emptying event times were manually recorded by a human observer. The dataset comprises 580986 samples of data with 64 bag-emptying events.

Each of the 3 data streams from the 3-axis accelerometer are filtered with a bandpass filter. The high-cut frequency is set to reduce aliasing. The low-cut frequency is set to remove any offsets caused by gravitational effects. The three filtered accelerometer data streams and the RSSI data stream are combined to form a dataset of four data streams, which are scaled to a range of $[0,1]$.

\subsection{Dataset Balancing}
\label{sec:balancedDataset}

The dataset is imbalanced where only $28\%$ of the samples are associated with bag-emptying events. This imbalance can create an undesirable bias in the classifiers. Resampling is used to form a balanced dataset. Sequences of samples are extracted to ensure the sequential nature of the data is maintained. To introduce some form of randomness in the sampling, the length of the extracted sequence is selected according to a normal distribution. The mean length $\mu$ and variance $\sigma$ of the bag-emptying event sequences are calculated from the sequence length of the manual label bag-emptying events. For each bag-emptying event sequence, only the $n \in \mathcal{N}(\mu, \sigma)$ preceding non-bag-emptying event samples are preserved. All remaining non-bag-emptying event samples are removed. The result is a dataset with sequence lengths that are similarly distributed between classes.


\section{Bag-Emptying Event Labelling}
\label{sec:bagDropLabelling}

The supervised machine learning algorithms require labels of the bag-emptying events for training. Although the times that the bag-emptying events occurred were recorded, the duration of the events were not recorded. Furthermore, owing to human error, the recoded bag-emptying times are only considered to be approximations. The bag-emptying event times are thus required to be refined and the bag-emptying event durations are required to be determined. For this, we consider two approaches: (1) a manual labelling approach using expert knowledge and (2) a semi-supervised approach based on K-means clustering.

\subsection{Manual Labelling of Bag-Emptying Events}
\label{sec:bagDropEvents}

To manually determine the bag-emptying event times and durations, the following bag-drop process is noted: (1) A pickers gait changes under the strain of a full bag as they walk from the trees to the bin; (2) the picker lifts the bag into the bin; (3) a flap at the bottom of the bag is opened to release the fruit (e.g. see \figurename{~\ref{fig:fruit_picker}}); (4) the bag is shaken and tugged to empty it; and (5) once the bag is empty, it is removed from the bin and the bottom flap is reattached.

\begin{figure}[!t]
	\centering
	\includegraphics[width=3.2in]{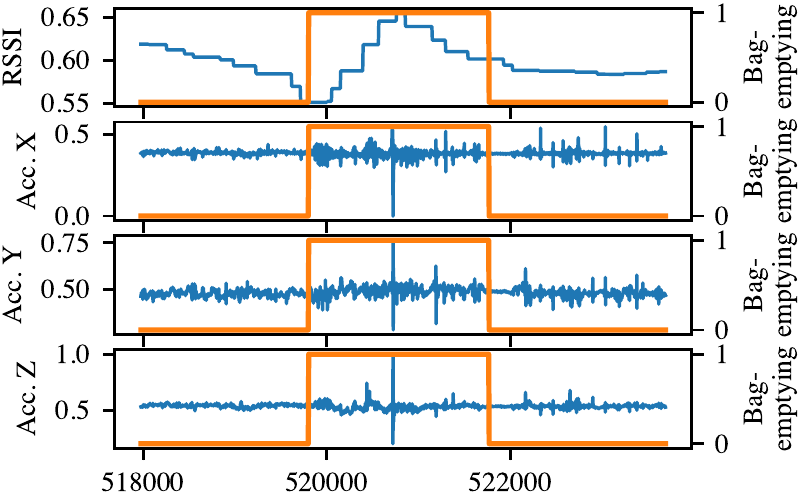}
	\caption{Plot of the sensor data with a bag-emptying event. Sensor data is plotted in blue. The manually defined bag-emptying event is plotted in orange with a value of 1 indicating a bad drop event.}
	\label{fig:bagdrop}
\end{figure}

The scaled accelerometer and RSSI data surrounding a bag-emptying event are illustrated in \figurename{~\ref{fig:bagdrop}}. The plot indicates a change in the dynamics as the picker transitions from normal picking activity to the bag-emptying event. The RSSI increases as the picker approaches the bin and the accelerometer signal level increases due to gait change. A spike in the accelerometer data occurs as the picker lifts the bag into the bin (this spike is not evident in every bag-emptying event). The accelerometer signal level remains high as the bag is shaken, removed and reassembled. The RSSI and the accelerometer signal decrease as the picker returns to fruit picking activities. Based on these observations the bag-emptying times were manually refined and the duration of the bag-emptying events were defined. The average bag-emptying event duration is 50 seconds with the emptying of the bag typically lasting between 10 and 20 seconds.
%

\subsection{Semi-supervised labelling of Bag-Emptying Events}

The K-means clustering algorithm is used to perform semi-supervised learning of the bag-emptying event labels\footnote{The K-means algorithm is selected due to its computational efficiency, however more complex clustering algorithms could also be considered.}. The approach is to pre-define the duration of bag-emptying events around the logged bag-emptying times. The K-means algorithm is then used to refine the duration by clustering samples according to statistical features of bag-emptying event and non-bag-emptying event data.

Data sample labels are initialised by defining the duration of all bag-emptying events to be 1200 samples in length, with 500 samples before and 700 samples after the manually logged bag-drop event time. Samples within this window are associated with bag-emptying events and samples outside this window are associated with normal fruit picking activity.

A 256-sample sliding window is shifted sample-by-sample over the data sequences. For each shift, a set of statistical features are extracted to produce a feature vector associated with each sample. This feature vector comprises the mean, standard deviation, minimum, maximum, and standard deviation of the derivative for the associated sample. Additionally, to capture the sequential structure of the data, an indicator is included to specify whether the neighbouring samples labels are associated with bag-emptying events. 

K-means algorithm is initialised by grouping the feature vectors into two clusters according to the initial bag-emptying event labels. The mean values of each cluster form the initial values for the K-means clustering algorithm. These clusters are refined using the K-means algorithm, which corresponds to refining the labels of each data sample. The algorithm was typically run over 10 iterations, where the number of iterations can affect how much the clusters change from the predefined settings.

The K-means clustering algorithm can produce false positives and false negatives, where false positives are spurious bag-emptying events predicted far from the logged time and false negatives are spurious non-bag-emptying events predicted near logged bag-emptying event times.

\begin{figure}[!t]
    \centering
    \includegraphics[width=3.2in]{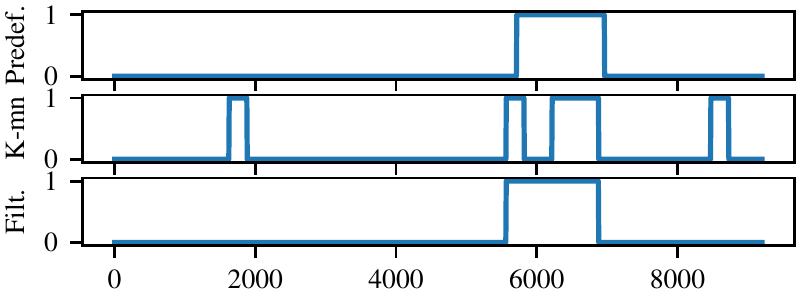}
    \caption{Plots of the process of semi-supervised bag-emptying event labelling. The predefined labels, the K-means clustering results, and the filtered K-means clustering results are plotted in the top, middle and bottom figures respectively.}
    \label{fig:auto_labelling}
\end{figure}

The results of the K-means algorithm are thus filtered using the intersection and union operators of mathematical set theory. Predicted bag-emptying event sequences are compared with the predefined bag-emptying event sequence as illustrated in \figurename{~\ref{fig:auto_labelling}}. To reduce false positives an intersection operator is used. If none of the samples in the predicted sequence overlap with the predefined sequence, the prediction is rejected. As illustrated in the bottom plot of \figurename{~\ref{fig:auto_labelling}}, the predictions at the beginning and end of the sequence are removed as they do not overlap with the predefined bag-emptying event sequence. To reduce false negatives, a union operation is applied between the predicted and predefined sequences. As illustrated in the bottom plot of \figurename{~\ref{fig:auto_labelling}}, the false negative is removed. Note that this operator is only applied to join predicted bag-emptying events. It is not permitted to extend the predicted bag-emptying event duration. This operation can however result in an unreasonably long bag-emptying event sequence if it joins two long sequences of positive samples. If a predicted sequence is longer than twice the predefined bag-emptying event sequence, it is rejected and the predefined bag-emptying event sequence is used instead.
%

The overall algorithm for the semi-supervised labelling approach is presented in Algorithm \ref{alg:unsupervisedLabeling}.
\begin{algorithm}[!t]
    \centering
    \caption{Semi-supervised Labelling of bag-emptying events using the K-means algorithm.}
    \label{alg:unsupervisedLabeling}
    \begin{algorithmic}[1]
        \REQUIRE The dataset $X$, a vector of bag-emptying event start times $t^{\text{start}}$, a vector of bag-emptying event end times $t^{\text{end}}$, and the dataset labels $Y$.
        \STATE Define the window size $q=256$
        \STATE Define an empty set of dataset labels $\hat{Y} = \varnothing$
        \FOR{each bag-emptying event, $i$}
        \STATE Extract dataset sequence surrounding the $i^\text{th}$ event \\
        \hspace{10pt} $A=X_{t^{\text{end}}_{i-1} : t^{\text{start}}_{i+1}}$
        \STATE Extract the labels associated with $A$ \\
        \hspace{10pt} $B = Y_{t^{\text{end}}_{i-1} : t^{\text{start}}_{i+1}}$
        \FOR{each index $j$ of sample in $A$}
        \STATE Extract the window of data associated with sample $j$\\
        \hspace{10pt} $W = A_{j:j+q}$
        \STATE Compute the feature vector for sample $j$ \\
        \hspace{10pt} $f_j = \text{features}(W)$
        \STATE Set the class of sample $j$ according to the window\\
        \hspace{10pt}
        $c_j =
        \begin{cases}
            0 & \frac{1}{q} \sum_{k=1}^q B_{j+k} < 0.5 \\
            1 & \frac{1}{q} \sum_{k=1}^q B_{j+k} \geq 0.5
        \end{cases}
        $
        \ENDFOR
        \STATE Update the classes using the K-means algorithm\\
        \hspace{10pt} $c \leftarrow \text{kmeans}(f, c)$
        \STATE Filter the labels using mathematical set theory \\
        \hspace{10pt} $c \leftarrow \text{filter}(c, B)$
        \STATE Append the filtered labels $c$ to the new label set $\hat{Y}$.
        \ENDFOR
        \RETURN The updated labels $\hat{Y}$.
    \end{algorithmic}
\end{algorithm}
%


\section{Models}
\label{sec:models}

Detecting a bag-emptying event from the wearable sensor data is a challenging task. As described in Section \ref{sec:bagDropLabelling}, the events comprises various sub-activities involved in the bag-emptying event. Furthermore, the bag-drop signals may vary according to pickers and the environment. The models are required to handle these variations.

\subsection{Feature Based Ensemble Model}

\begin{figure}[!t]
	\centering

\def\horisep{0.90}
\def\vertsep{0.2}

\definecolor{mygray}{gray}{0.4}

\begin{footnotesize}
\begin{tikzpicture}[thick, draw=black!45,  >={Latex[length=1.5mm, width=1.2mm, black!45]}]

	\node[draw=black!70, fill=white, minimum width=40pt, minimum height=20pt, label=below:RSSI](rssi) 
	at (126pt, 0pt) {};
	\draw[very thick, black!30, smooth, domain=-19:19] plot(\x + 126pt, 5*rand + 0pt);

	\node[draw=black!70, fill=white, minimum width=40pt, minimum height=20pt, label=below:Acc. Z](accz) 
	at (84pt, 0pt) {};
	\draw[very thick, black!30, smooth, domain=-19:19] plot(\x + 84pt, 5*rand + 0pt);

	\node[draw=black!70, fill=white, minimum width=40pt, minimum height=20pt, label=below:Acc. Y](accy) 
	at (42pt, 0pt) {};
	\draw[very thick, black!30, smooth, domain=-19:19] plot(\x + 42pt, 5*rand + 0pt);

	\node[draw=black!70, fill=white, minimum width=40pt, minimum height=20pt, label=below:Acc. X](accx) 
	at (0pt, 0pt) {};
	\draw[very thick, black!30, smooth, domain=-19:19] plot(\x pt, 5*rand + 0pt);

	\node[draw=black!70, minimum width=100pt, minimum height=30pt,align=center](features) 
	at (63pt, 40pt) 
	{Feature vector \\
		\textcolor{mygray}{(std., energy, RMS($dx / dt$), RMS($d^{2}x / dt^{2}$),} \\
			\textcolor{mygray}{mean($dx / dt$), mean($d^{2}x/dt^{2}$), min, max)}};
	\path[->] (rssi.north) edge (features);
	\path[->] (accz.north) edge (features);
	\path[->] (accy.north) edge (features);
	\path[->] (accx.north) edge (features);
	
	\node[draw=black!70, minimum width=50pt, minimum height=15pt,align=center](NB) 
	at (33pt, 75pt) {Naive Bayes};
	\path[->] (features) edge (NB.south);
	
	\node[draw=black!70, minimum width=50pt, minimum height=15pt,align=center](ANN) 
	at (93pt, 75pt) {ANN \textcolor{mygray}{(512)}};
	\path[->] (features) edge (ANN.south);
	
	\node[circle, draw=black!70, fill=black!2, minimum height=15pt,align=center](sum) 
	at (63pt, 110pt) {$\sum$};
	\path[->] (NB.north) edge node[left] {$60\%$} (sum);
	\path[->] (ANN.north) edge node[right] {$40\%$} (sum);	
	
	\node[circle, draw=black!70, fill=black!5, minimum height=10pt,align=center](class) 
	at (63pt, 135pt) {};
	\path[->] (sum) edge (class);
	
	\draw [-, decorate,decoration={brace, amplitude=5pt,raise=0pt},black] ((150pt,15pt) -- (150pt,-20pt) node [black,midway,xshift=5pt, right, align=left] {Input \\data};

	\draw [-, decorate,decoration={brace, amplitude=5pt,raise=0pt},black] ((150pt,60pt) -- (150pt,22pt) node [black,midway,xshift=5pt, right, align=left] {Feature \\vector};

	\draw [-, decorate,decoration={brace, amplitude=5pt,raise=0pt},black] ((150pt,120pt) -- (150pt,65pt) node [black,midway,xshift=5pt, right] {Ensemble};
	
	\draw [-, decorate,decoration={brace, amplitude=5pt,raise=0pt},black] ((150pt,140pt) -- (150pt,125pt) node [black,midway,xshift=5pt, right] {Class};

\end{tikzpicture}
\end{footnotesize}
	\caption{Architecture of the ensemble model. A feature vector is assembled from a window of data. A naive Bayes and an ANN perform a classification given the feature vector. The outputs of these classifiers are weighted and summed to determine the class.}
	\label{fig:ensemble_model}
\end{figure}
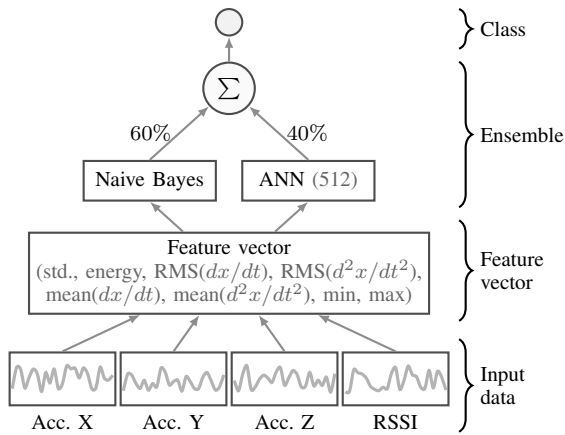
A traditional feature-based ensemble model is applied for detecting bag-emptying events. The model is illustrated in \figurename{~\ref{fig:ensemble_model}}. A 256 sample sliding window with zero overlap is applied to the data. The following features are extracted in each window: standard deviation, energy, the RMS first and second derivative, mean first and second derivative, minimum value, and maximum value \cite{Ravi2005Activity, Atallah2011Sensor}. 

The features are provided as inputs to an ensemble classifier comprising a Gaussian Naive Bayes classifier and a neural network. These are two commonly used classifiers in HAR \cite{Lara2013Survey} and are sufficiently different from each other to provide diversity in the ensemble. The neural network comprises a single hidden layer with 512 neurons with hyperbolic tangent activation functions. The ADAM algorithm \cite{kingma2014adam} is used to train the neural network. The parameters of the Gaussian naive Bayes classifier are estimated using maximum likelihood. Both classifiers output a prediction in the form of a probability of a bag-emptying event. These predictions are combined in the ensemble through a weighted summation. The Naive Bayes classifier is weighted with 60\% of the vote (which was determined through cross validation).

\subsection{Recurrent Convolutional Neural Network (RCNN)}

\begin{figure}[!t]
	\centering

\def\horisep{0.90}
\def\vertsep{0.9}

\definecolor{mygray}{gray}{0.4}

\begin{footnotesize}
\begin{tikzpicture}[thick, draw=black!45,  >={Latex[length=1.5mm, width=1.2mm, black!45]}]

	\node[draw=black!70, fill=white, minimum width=40pt, minimum height=20pt, label=below:RSSI](rssi) at (108pt, 12pt) {};
	\draw[very thick, black!30, smooth, domain=-19:19] plot(\x + 108pt, 5*rand + 12pt);

	\node[draw=black!70, fill=white, minimum width=40pt, minimum height=20pt, label=below:Acc. Z](accz) at (72pt, 8pt) {};
	\draw[very thick, black!30, smooth, domain=-19:19] plot(\x + 72pt, 5*rand + 8pt);

	\node[draw=black!70, fill=white, minimum width=40pt, minimum height=20pt, label=below:Acc. Y](accy) at (36pt, 4pt) {};
	\draw[very thick, black!30, smooth, domain=-19:19] plot(\x + 36pt, 5*rand + 4pt);

	\node[draw=black!70, fill=white, minimum width=40pt, minimum height=20pt, label=below:Acc. X](accx) at (0pt, 0pt) {};
	\draw[very thick, black!30, smooth, domain=-19:19] plot(\x pt, 5*rand + 0pt);

	\node[draw=black!70, minimum width=160pt, minimum height=50pt](inputs) at (54pt, 0pt) {};
	
	\node[draw=black!70, minimum width=150pt, minimum height=15pt](conv1) at (54pt, 44pt) 
	{Conv. 1 \textcolor{mygray}{(filters = 32, kernel size = 32)}};
	\path[->] (inputs) edge (conv1);
	
	\node[draw=black!70, minimum width=130pt, minimum height=15pt](pool1) at (54pt, 70pt) 
	{Pool 1 \textcolor{mygray}{(pool size = 8, stride = 4)}};
	\path[->] (conv1) edge (pool1);
	
	\node[draw=black!70, minimum width=150pt, minimum height=15pt](conv2) at (54pt, 100pt) 
	{Conv. 2 \textcolor{mygray}{(filters = 16, kernel size = 16)}};
	\path[->] (pool1) edge (conv2);
	
	\node[draw=black!70, minimum width=130pt, minimum height=15pt](pool2) at (54pt, 130pt) 
	{Pool 2 \textcolor{mygray}{(pool size = 4, stride = 2)}};
	\path[->] (conv2) edge (pool2);
	
	\node[draw=black!70, minimum width=100pt, minimum height=15pt](dense) at (54pt, 160pt) 
	{Dense \textcolor{mygray}{(512 neurons)}};
	\path[->] (pool2) edge (dense);

	\node[draw=black!70, fill=black!15, minimum width=18pt, minimum height=18pt](rnn1) at (-10pt, 190pt) {};
	\node[draw=black!70, fill=black!15, minimum width=18pt, minimum height=18pt](rnn2) at (54pt, 190pt) {};
	\node[draw=black!70, fill=black!15, minimum width=18pt, minimum height=18pt](rnn3) at (120pt, 190pt) {};	
	
	\draw[->] (dense) edge (rnn2);
	\draw[->] (rnn1) edge (rnn2);	
	\draw[->] (rnn2) edge (rnn3);	
	
	\node[circle, draw=black!70, fill=black!5, minimum width=10pt, minimum height=10pt](out1) at (-10pt, 220pt) {};
	\node[circle, draw=black!70, fill=black!5, minimum width=10pt, minimum height=10pt](out2) at (54pt, 220pt) {};
	\node[circle, draw=black!70, fill=black!5, minimum width=10pt, minimum height=10pt](out3) at (120pt, 220pt) {};
	\draw[->] (rnn1) edge (out1);
	\draw[->] (rnn2) edge (out2);
	\draw[->] (rnn3) edge (out3);
	
%
	
	\draw [-, decorate,decoration={brace, amplitude=5pt,raise=0pt},black] ((140pt,30pt) -- (140pt,-30pt) node [black,midway, xshift=20pt, align=left] {Input \\tensor};
	\draw [-, decorate,decoration={brace, amplitude=5pt,raise=0pt},black] ((140pt,170pt) -- (140pt,35pt) node [black,midway, xshift=18pt, align=left] {CNN};
	
%
	
	\draw [-, decorate,decoration={brace, amplitude=5pt,raise=0pt},black] ((140pt,205pt) -- (140pt,175pt) node [black,midway, xshift=18pt, align=left] {RNN};
	
	\draw [-, decorate,decoration={brace, amplitude=5pt,raise=0pt},black] ((140pt,230pt) -- (140pt,210pt) node [black,midway, xshift=22pt, align=left] {Outputs};

\end{tikzpicture}
\end{footnotesize}
	\caption{Architecture of the RCNN model for a single sequence sample (based on \cite{Ordonez2016Deep} and \cite{Yao2017DeepSense}). The input contains a tensor comprising a window of samples from each data stream. The CNN comprises two convolutional layers, two pooling layers, and a densely (fully) connected neural network. The CNN output is passed to an LSTM cell of the RNN. The LSTM cell output is passed to a sigmoidal neuron which outputs the class of the input window sample. The LSTM models temporal dynamics over sample windows.}
	\label{fig:model}
\end{figure}
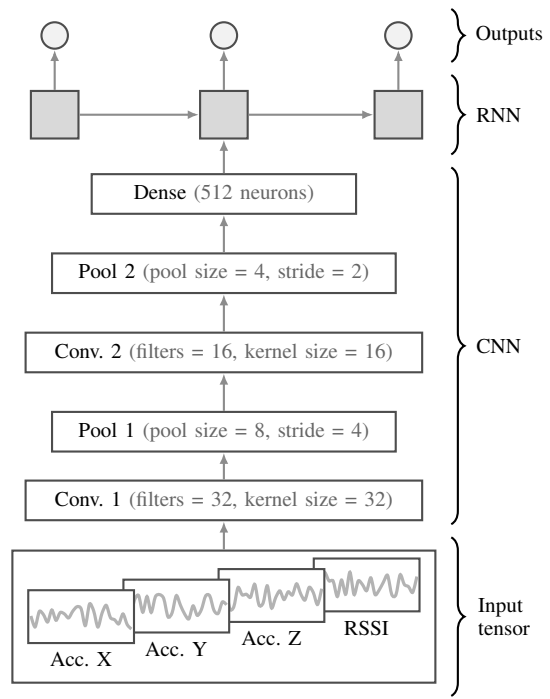

The architecture of the RCNN used in this study is illustrated in \figurename{~\ref{fig:model}} and is based on the models presented in \cite{Ordonez2016Deep} and \cite{Yao2017DeepSense}. For each data stream, a set of samples are extracted using a 256 sample sliding window with zero overlap. The windows of samples from the 4 sensor streams are combined into a tensor, where each stream represents channel for the input of the CNN portion of the model. A key advantage of the RCNN over the ensemble model is that it is an end-to-end model and does not require feature engineering.

The CNN portion of the RCNN performs feature extraction and the RNN portion of the RCNN models temporal dynamics over the sample windows. The RCNN outputs the class of the input window sample. The rectified linear unit (ReLU) is used as the activation function in hidden layers, the filter and layer sizes in the network were determined through trial and error, and the ADAM algorithm \cite{kingma2014adam} is used for training.

\section{Methodology}
\label{sec:methodology}


Both the models use a sliding window with zero overlap. The window of data is presented to the classifier. The set of data in the window is classified to belong either to a bag-emptying event or a non-bag-drop event. Each data sample within the window is associated with this predicted class. During training, a window may be slid to a position where it partially falls within a bag-emptying event. The ground-truth class label for the window is calculated as the average true class of all the samples in the window.


To validate models and results, a 6-fold cross validation test is performed. The dataset is split into 6 equal data segments. Each data segment is a continuous time series of 96831 samples. The model is trained on 5 of the data segments and tested on the remaining segment. This is repeated such that the model is tested on each data segment of the dataset.


Accuracy, precision, recall, and F-score are used to measure the performance of the classifiers presented in this study. Accuracy describes the ratio of the number of correct classifications to the total number of classified samples. Precision describes the ratio of correct classifications to the total number of classifications made for the particular class. It considers the number of incorrectly predicted bag-emptying events and thus provides a measure of the classifier quality. Recall describes the ratio of correct classifications to the total number of items which truly belong to the predicted class. It considers the number of bag-emptying event samples that were missed. Recall thus provides a measure of the probability of correctly classifying the bag-emptying event. Finally, the F-score is defined as the harmonic mean between the precision and recall. It provides a measure to describe both the precision and recall together.


\section{Results}
\label{sec:results}

\subsection{Comparison of Labelling Approaches}
	
The median value of the accuracy, precision, recall, and F-score across the 6-fold cross validations for the predefined labels, manually defined labels, and the learned labels are provided in Table \ref{table:medianResults}.
\begin{table}[t]
    \centering
	\caption{Feature-based ensemble model and RCNN model median value results for the predefined labels, manually defined labels, and the learned labels.}
	\label{table:medianResults}
	\begin{footnotesize}
	\begin{tabular}{llccc}
		\toprule
		Model & Measure & Predefined & Manual & Learned \\
		\midrule
		\multirow{4}{*}{Ensemble}
		& Accuracy 		& 74\% & \textbf{80}\%	& \textbf{80}\% \\
		& Precision 	& 81\% & \textbf{89}\%	& 83\% \\
		& Recall 		& 63\% & 66\% 			& \textbf{71}\% \\
		& F-score 		& 71\% & \textbf{77}\%	& \textbf{77}\% \\
		\midrule
		\multirow{4}{*}{RCNN}
		& Accuracy 		& 79\% 			& 76\% 	& \textbf{86}\% \\
		& Precision 	& 86\% 			& 80\% 	& \textbf{89}\% \\
		& Recall 		& 71\% 			& 72\%	& \textbf{83}\% \\
		& F-score 		& 77\% 			& 75\% 	& \textbf{84}\% \\
		\bottomrule
	\end{tabular}
	\end{footnotesize}
\end{table}

The ensemble model produces the poorest results with the predefined labels. Improved performance is obtained with the manually and learned labels. The performance for the manually and learned labels are similar. This is a key result as it validates the proposed approach to learning the dataset labels. The manual labels produce a higher precision than the learned labels. However, the learned labels provide a higher recall than the manual labels. The recall is considered to a more important measure as it relates to the probability of detecting bag-emptying events.

\begin{figure}[!t]
	\centering
    \subfloat[Ensemble model.]{\includegraphics[width=3.0in]{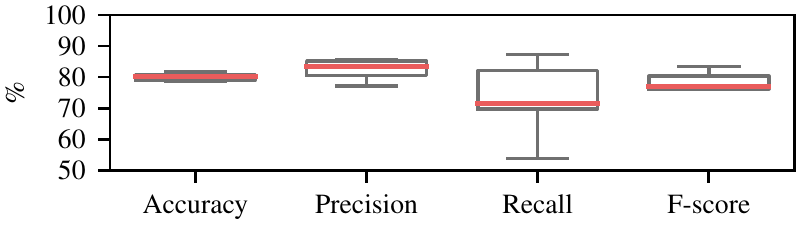}}
    \label{fig:ensemble_boxplot}
    \subfloat[RCNN model.]{\includegraphics[width=3.0in]{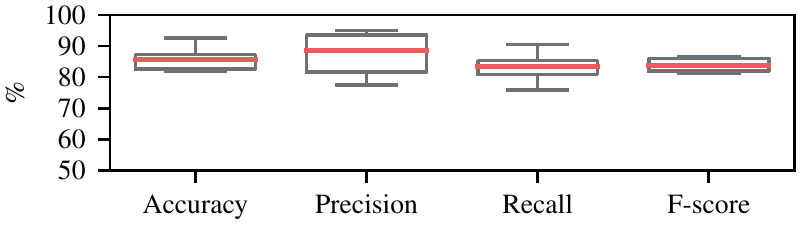}}
    \label{fig:conv_rnn_boxplot}
	\caption{Box whisker plot of the cross validation results.}
	\label{fig:boxPlot}
\end{figure}

The best results for the RCNN are obtained with the learned labels. This reinforces the validation of the semi-supervised label learning approach. The RCNN produces results that are much higher than those produced by the ensemble model. The ensemble model's precision for the manual labels matches the capability of the RCNN. This however is achieved at the cost of a low recall value of 66\%. Along with the high precision value, the RCNN produces a recall of 83\%. These comparisons are reiterated by the 7\% difference in F-score results between the two models.

\subsection{Results with the Learned Labels}

A box whisker plot of the results over the 6-fold cross validation test are presented in \figurename{~\ref{fig:boxPlot}}. The RCNN performs better than the ensemble model for all performance measures. The precision box of the RCNN reaches high values. However, the box is the largest of all boxes in the diagram. A larger box indicates that there is more variability in the precision results. The quality of the model is however still high considering that the range of the box remains above 80\%. The ensemble model's recall box is large with whiskers that extend to low values. This indicates a high level of uncertainty in the ensemble models recall results. The recall box of the RCNN is narrow indicating high certainty in the RCNN recall results. The range of the box remains above 80\% indicating good prediction ability of the RCNN model. Both models produce narrow F-score boxes. The range of the RCNN F-score box is higher than the ensemble model indicating superior performance overall.

\begin{figure*}[!t]
	\centering
	\subfloat[Ensemble model.]{\includegraphics[width=3.2in]{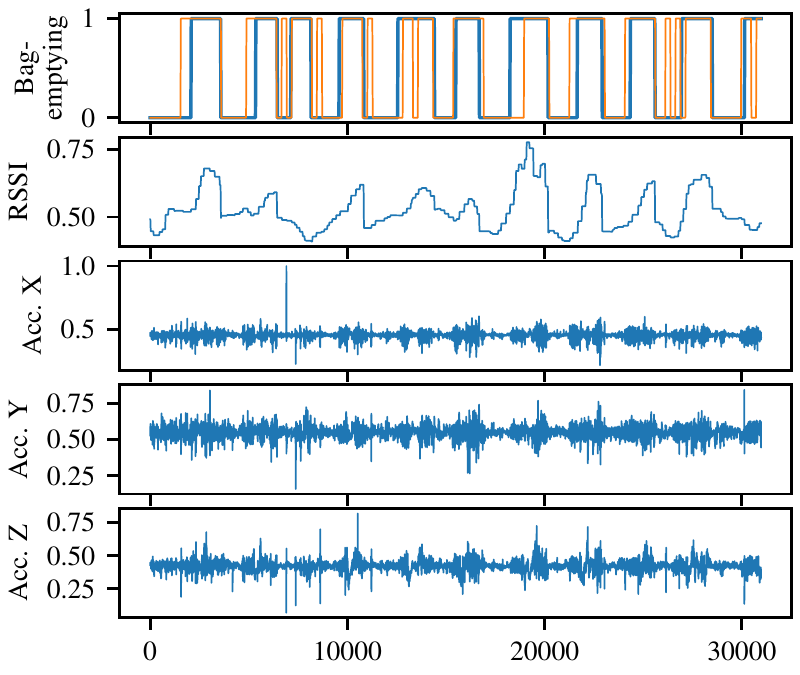}}
	\label{fig:ensemble_data_prediction_0}
	\subfloat[RCNN model.]{\includegraphics[width=3.2in]{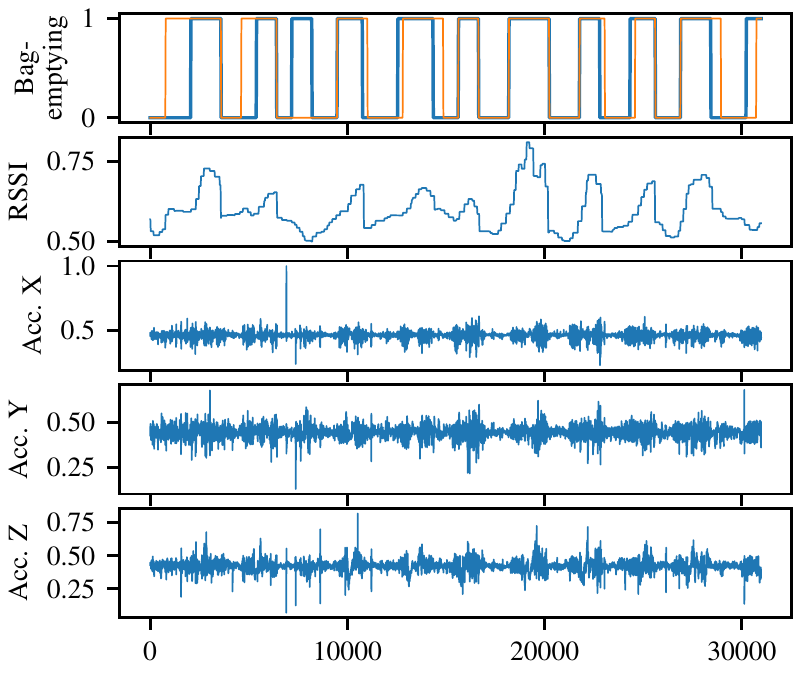}}
	\label{fig:conv_rnn_data_prediction_0}
	\caption{bag-emptying event predictions and dataset for the convolutional recurrent neural network. The blue curves plot the dataset. The orange curve the top figure plots the model predictions. A value of 1 indicates a bag-emptying event.}
	\label{fig:data_prediction}
\end{figure*}

A plot of the predictions and data for one of the cross validation folds is presented in \figurename{~\ref{fig:data_prediction}}. The learned labels correspond well with the changes in the accelerometer and RSSI data. The predictions of the ensemble model are sporadic resulting in several false positives and false negatives. The RCNN model predictions are smoother over time.  The RCNN however misses the third bag-emptying event. This is possibly due to the RSSI level unexpectedly dropping to a minimum during this bag-emptying event. Unlike the RCNN, the ensemble model is able to detect the third bag-emptying event. This seems to indicate that the RCNN relies more on RSSI data and the ensemble model relies more on accelerometer data. Note that the last bag-emptying event is not missed by the RCNN. It is detected in the following cross validation fold.

\begin{figure*}[!t]
    \captionsetup[subfigure]{labelformat=empty}
    \centering
    \subfloat[]{\includegraphics[width=3.2in]{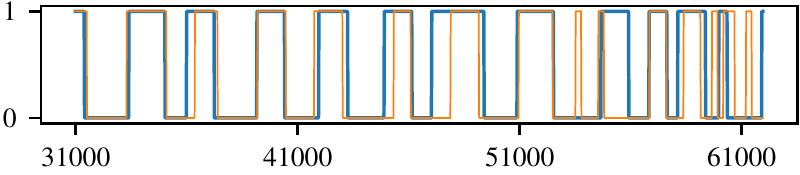}}
    \label{fig:ensemble_prediction_1}
    \hspace{10pt}
    \subfloat[]{\includegraphics[width=3.2in]{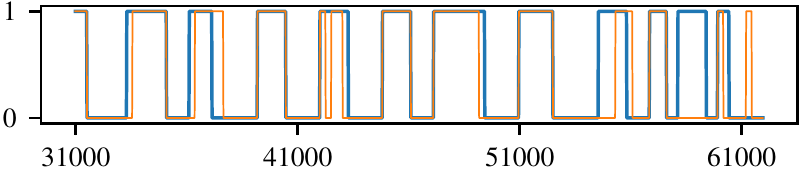}}
    \label{fig:conv_rnn_prediction_1}
    \\
    \vspace{-25pt}
    \subfloat[]{\includegraphics[width=3.2in]{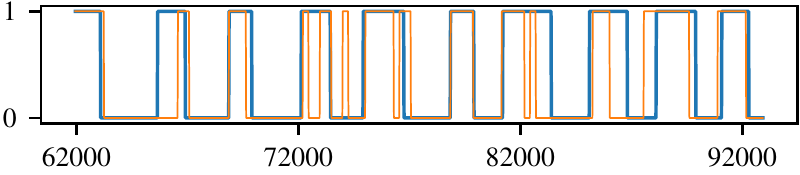}}
    \label{fig:ensemble_prediction_2}
    \hspace{10pt}
    \subfloat[]{\includegraphics[width=3.2in]{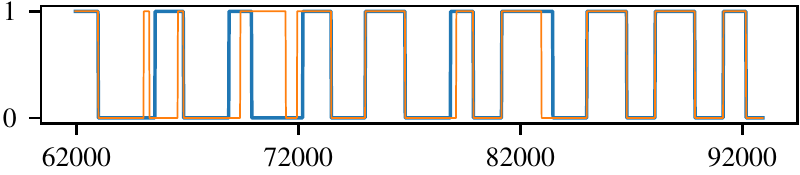}}
    \label{fig:conv_rnn_prediction_2}
    \\
    \vspace{-25pt}
    \subfloat[]{\includegraphics[width=3.2in]{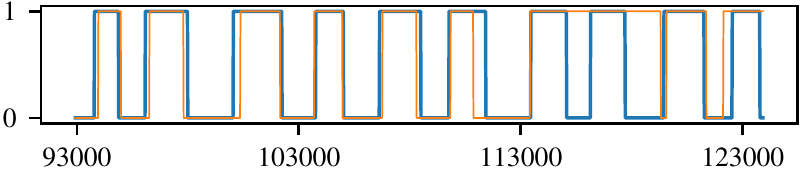}}
    \label{fig:ensemble_prediction_3}
    \hspace{10pt}
    \subfloat[]{\includegraphics[width=3.2in]{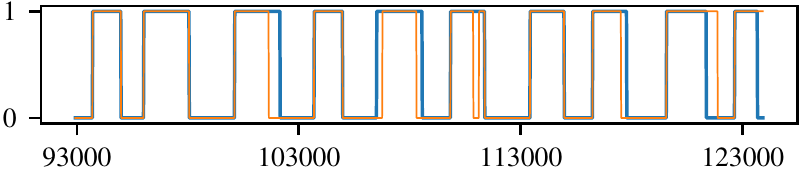}}
    \label{fig:conv_rnn_prediction_3}
    \\
    \vspace{-25pt}
    \subfloat[]{\includegraphics[width=3.2in]{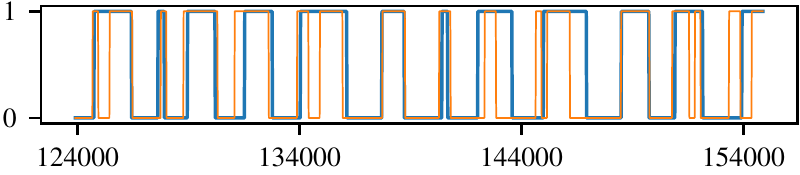}}
    \label{fig:ensemble_prediction_4}
    \hspace{10pt}
    \subfloat[]{\includegraphics[width=3.2in]{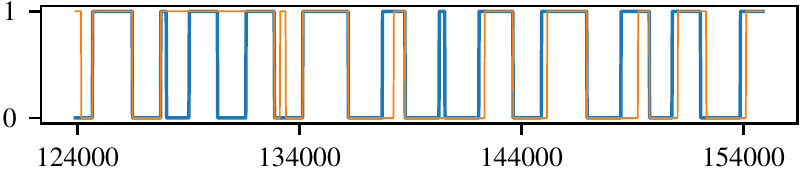}}
    \label{fig:conv_rnn_prediction_4}
    \\
    \vspace{-25pt}
    \subfloat[]{\includegraphics[width=3.2in]{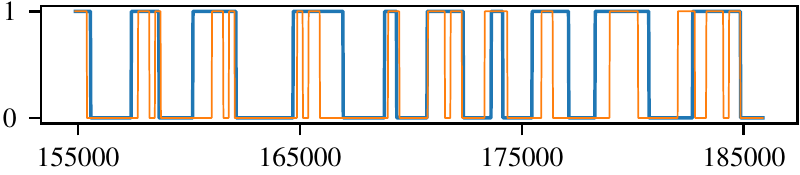}}
    \label{fig:ensemble_prediction_5}
    \hspace{10pt}
    \subfloat[]{\includegraphics[width=3.2in]{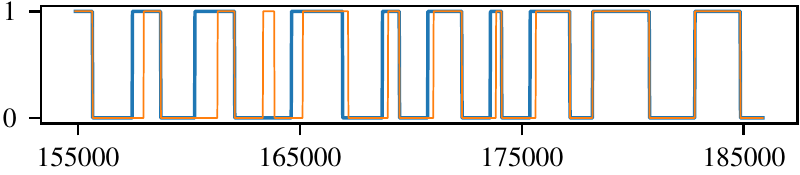}}
    \label{fig:conv_rnn_prediction_5}
    \caption{Bag-emptying event predictions for the second to sixth cross validation folds along the rows. Plots of the ensemble model are presented in the left column and plots of the RCNN model are presented in the right column. The horizontal axis is the sample number. The vertical axis is the probability of a bag-emptying event. The blue curve plots the manual bag-emptying events and orange curve plots the predicted bag-emptying events.}
    \label{fig:predictions}
\end{figure*}

The plots of the predictions for the remaining cross validation folds are illustrated in \figurename{~\ref{fig:predictions}}. The ensemble model results are more sporadic and the predictions are more confident. The RCNN is less confident and the predictions are smoother over time. This is preferred when false positives have a high risk. 
%
%


\section{Discussion and Conclusion}
\label{sec:conclusion}

In this study, we present a novel application on measuring fruit picker productivity. The picker productivity is measured by detecting bag-emptying events from wearable sensor data. A traditional feature-based ensemble model and a deep convolutional recurrent neural network are applied to predict bag-emptying events from the wearable sensor data. Furthermore, a semi-supervised method for learning the bag-emptying event labels is presented.

Results indicate that both models are able to successfully detect the bag-emptying events. The RCNN model is more accurate than the ensemble model but it is less confident, which results in 3 of the 64 bag-emptying events being missed. The ensemble model has at least one positive detection within each bag-emptying event suggesting that all bag-emptying events. As the ensemble model does not directly model any temporal relationship between samples, its predictions are noisy resulting in multiple detections for a single bag-emptying event. The RCNN models temporal dynamics to produces smooth predictions over time and more accurate predictions of bag-emptying event durations.

In future work, the RCNN could be improved by increasing its capacity and training it on more data. The capacity of the RCNN can be increased by introducing more CNN filters and by increasing the CNN depth. Such improvements may provide the model with the capability to learn more advanced features. The RNN can be improved by adding multiple layers and by using bi-directional RNNs. More complex models however generally require more data. Collecting more data is thus a priority for future work. Other than improving models, the data captured by the sensors provide information relating to other problems such as health and safety. For example, the accelerometer sensor can be used to detect excessive bag weight or falls.


    \bibliographystyle{IEEEtran}
    \bibliography{IEEEabrv,bibliography}
    
\end{document}